\documentclass[10pt,twocolumn,letterpaper]{article}

\usepackage{cvpr}              % To produce the CAMERA-READY version
\usepackage{graphicx}
\usepackage{amsmath}
\usepackage{amssymb}
\usepackage{booktabs}
\usepackage{pifont}
\usepackage{adjustbox}
\usepackage{verbatim}
\usepackage{multirow}
\usepackage{tabu}
\usepackage{siunitx}
\usepackage{array}
\usepackage{lipsum}

\usepackage[pagebackref,breaklinks,colorlinks]{hyperref}
\usepackage[capitalize]{cleveref}

\usepackage[pagebackref,breaklinks,colorlinks]{hyperref}

\usepackage[capitalize]{cleveref}
\crefname{section}{Sec.}{Secs.}
\Crefname{section}{Section}{Sections}
\Crefname{table}{Table}{Tables}
\crefname{table}{Tab.}{Tabs.}

\def\cvprPaperID{*****} % *** Enter the CVPR Paper ID here
\def\confName{CVPR}
\def\confYear{2022}

\begin{document}

%\title{Improved Edge Segmentation}
\title{Efficient Heterogeneous Video Segmentation at the Edge}

\author{
Jamie Menjay Lin~~~
Siargey Pisarchyk~~~
Juhyun Lee~~~
David Tian~~~
Tingbo Hou~~~\\
Karthik Raveendran~~~
Raman Sarokin~~~
George Sung~~~
Trent Tolley~~~
Matthias Grundmann~~~\\
Google, Mountain View, CA, USA\\
%1600 Amphitheatre Pkwy, Mountain View, CA 94043, USA\\
{\tt\small \{jmlin,siargey,impjdi,dctian,krav,tingbo,sorokin,gsung,trenttolley,grundman\}@google.com}
%\and
}
\maketitle

%%%%%%%%% ABSTRACT
\begin{abstract}
We introduce an efficient video segmentation system for resource-limited edge devices leveraging heterogeneous compute. Specifically, we design network models by searching across multiple dimensions of specifications for the neural architectures and operations on top of already light-weight backbones, targeting commercially available edge inference engines. We further analyze and optimize the heterogeneous data flows in our systems across the CPU, the GPU and the NPU. Our approach has empirically factored well into our real-time AR system, enabling remarkably higher accuracy with quadrupled effective resolutions, yet at much shorter end-to-end latency, much higher frame rate, and even lower power consumption on edge platforms.
\end{abstract}
\vspace{-6mm}

%%%%%%%%% BODY TEXT
\section{Introduction}
\label{sec:introduction}

Video segmentation is a foundational technology powering various computer vision tasks in business and entertainment use cases, such as video editing in augmented reality (AR) and background blur and replacement in video conferencing.
Machine learning (ML) inference for high-quality real-time video segmentation has been a challenging problem particularly on resource-limited edge devices,~\eg smartphones, with model accuracy, real-time latency, and power consumption being the key areas for improvement.

\begin{figure}[t]
\centering
\includegraphics[width=\columnwidth]{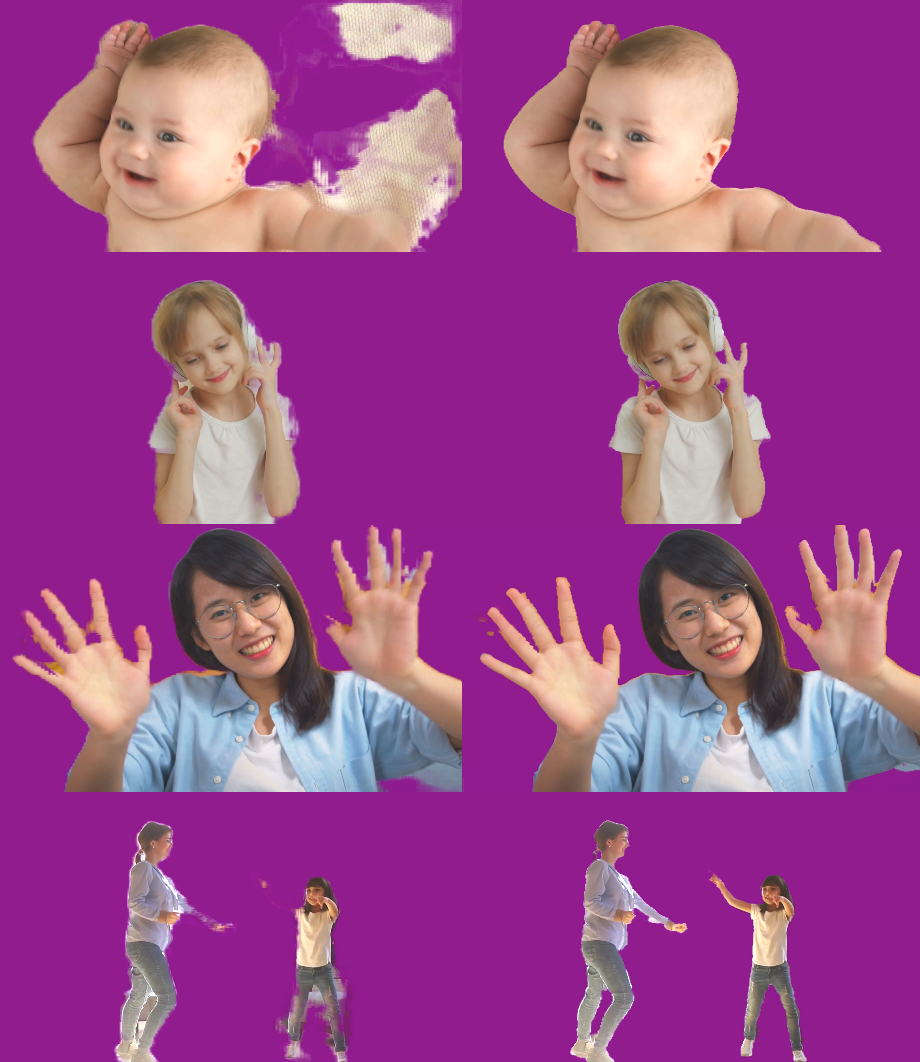}
\caption{Real-time video segmentation comparison on Google Pixel 6 smartphone. Left column: Baseline ML inference at $89.5\%$ mIoU~\cite{iou} (segmentation quality metric) and 2.03 Watts. Right column: Our ML inference at $95.1\%$ mIoU and 1.84 Watts.}
\label{fig:qualitative}
\vspace{-8mm}
\end{figure}

We present practical techniques for real-time high-quality video segmentation at the edge (\Cref{fig:qualitative}). We describe i) our advances made to common light-weight network architectures,~\eg MobileNetV3~\cite{mobilenetv3} and EfficientNetLite~\cite{efficientnet-lite}, and ii) our optimizations to the inference pipelines at the edge, such as WebGL-based~\cite{webgl} browsers and the NPU inference pipelines on mobile devices. 

\section{Related Work}
\label{sec:relatedwork}

In the course of neural network-based model development for semantic segmentation, fully convolutional networks~\cite{7478072} were introduced as an earlier progress. Numerous follow-up works continued to improve on accuracy and efficiency, proposing various types of model architectures for segmentation, such as DeepLab~\cite{7913730}, Vortex Pooling~\cite{8995234}, PSPNet~\cite{zhao2017pspnet}, and HRNet~\cite{8953615, 9578746}. Another recent line of work targets the problem of image matting for foreground-background separation, which typically requires either the availability of a trimap~\cite{ifm, 9578641, 9578248} or a pre-captured background image~\cite{ifm, BMSengupta20, 9578641}. Dynamic routing~\cite{li2020learning} was introduced on a generalized view over various architectures for segmentation. Neural architecture search techniques have been another line of work in searching for optimized segmentation model architectures~\cite{NEURIPS2018_c90070e1, 8954247, 8953530}, as well as particular types of model improvements such as transformer-based segmentation models~\cite{9578324, 9564969}.
% Our video segmentation is lightweight, not requiring a trimap nor a pre-captured background image during inference, and end-to-end optimized with neural architecture search on hardware accelerators such as GPUs and NPUs~\cite{edge-tpu,hexagon}.

\section{Edge Segmentation Model Design}
\label{sec:model}
% \subsection{Practical Considerations}
Neural networks running on edge devices are often tailored to the available computational resources. As a baseline for the segmentation task, we consider a commonly used fully convolutional U-net architecture based on a MobileNetV3 backbone.
Running CPU model inference with an input resolution of $256\hspace{-1mm}\times\hspace{-1mm}256$ takes
% 20ms on an Intel Core i9 processor ~\cite{inteli9}, and
over 50ms on an Intel Celeron N3060 as used in common EDU-targeted Chromebooks, suggesting that even a low-quality segmentation model is too expensive to run at real-time on older hardware.

To allow for larger models while still supporting real-time video, we targeted GPU and NPU that can
deliver the required FLOPS~\cite{flops} or TOPS~\cite{tops} for the task of high resolution video segmentation.
% With the required FLOPs for the task,
% we conclude to leverage hardware acceleration,~\eg the GPU and the NPU, and focus on designing a segmentation model
% that is compatible with such accelerators.
We analyze the impact of our design choices including the structure of the encoder and decoder blocks, convolution types, input resolutions, and width and depth multipliers.
\Cref{table:ablation} summarizes the full ablation study discussed in this section.

\begin{figure}[t]
\centering
\includegraphics[width=\columnwidth]{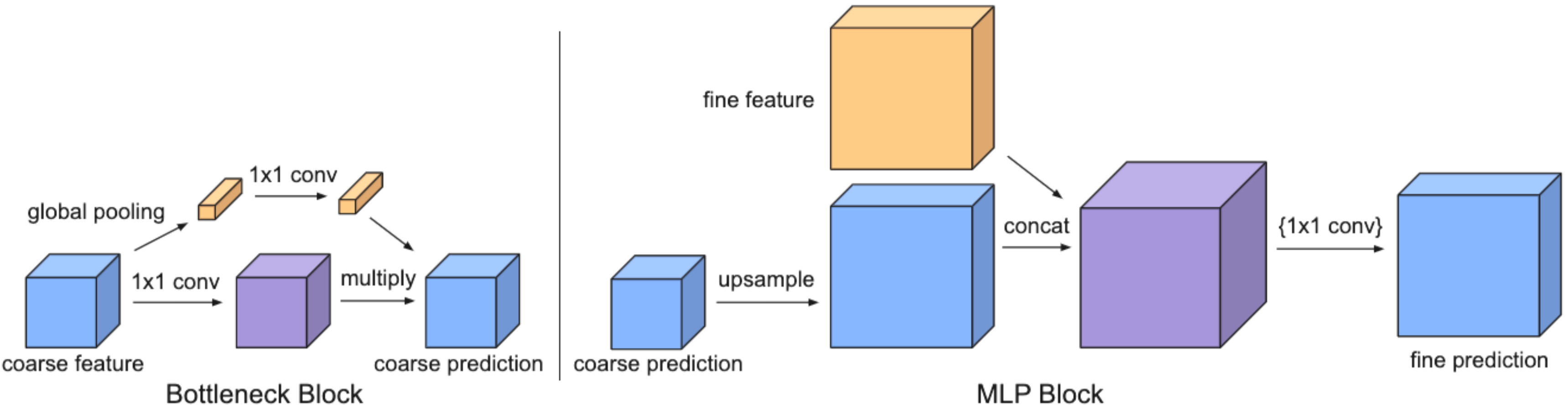}
\caption{Left: The bottleneck block paralleled with a squeeze-and-excitation operation, which increases GPU latency due in part to the global pooling operations. Right: Our decoder with the MLP operation, which does not involve global pooling operations.}
\label{fig:decoder}
\end{figure}

% \subsection{Model Architecture}
While the squeeze-and-excitation blocks\cite{squeeze} in MobileNetV3 are efficient on the CPU, they increase the latency on GPUs due to the global average pooling of large tensors. We re-base our architecture on EfficientNetLite where each encoder block comes with an inverted residual bottleneck and produces an output tensor that is $1/32$ of the input. The decoder then begins with a bottleneck block that predicts a coarse segmentation mask as shown in \cref{fig:decoder}. In this instance, the global average pooling only needs to operate on a much smaller tensor and the prediction accuracy benefits from such channel attention mechanism.
%increases the accuracy of the prediction at a marginally higher latency compared to the standard implementation as can be seen in Table \ref{table:ablation}.

Next, we explore three variants for the decoder design:
\begin{enumerate}
\item \textbf{Bilinear Upsampling}: Upsamples bilinearly  the lower resolution tensor to match the resolution of the next higher encoder block. This applies a non-linearity and sum with the tensor from the encoder skip connection.
\item \textbf{Channel Attention}: Applies squeeze-and-excitation blocks after upsampling, followed by a set of $1\hspace{-1mm}\times\hspace{-1mm}1$ convolutions and a $3\hspace{-1mm}\times\hspace{-1mm}3$ depth-wise convolution. 
\item \textbf{Multilayer Perceptron (MLP)}: Applies upsampling followed by a sequence of $1\hspace{-1mm}\times\hspace{-1mm}1$ convolutions (MLPs). 
\end{enumerate}

We found that $1\hspace{-1mm}\times\hspace{-1mm}1$ convolutions make a significant difference to the accuracy of the model when it comes to finer details, and that the MLP block serves as a reasonable performance compromise compared to the more expensive attention block.  We also considered the effect of various convolution types on the network, ranging from standard convolutions to other types of convolutions. We found while separable convolutions are a natural fit for modern GPUs, certain hardware such as NPUs can benefit from group convolutions due to increased chip utilization, enhancing the model expressivity without increasing the latency.  Finally, we performed a hyperparameter search for the EfficientNetLite family and settled on a multiplier of $1.0$.

Having settled on an architecture, we measured the trade-offs between input resolution, latency, and accuracy. In each of these cases, we trained the network using images down-scaled to the specified input resolution and measured the IoU by down-scaling the ground truth mask to the output resolution. We found that resolution has the largest impact on the quality of segmentation, but observed rapidly diminishing returns to increasing the resolution beyond $640\hspace{-1mm}\times\hspace{-1mm}640$.

\begin{table}[t]
\centering
\begin{adjustbox}{width=1.00\linewidth}
\begin{tabular}{|c|c|c|c|c|c|c|}
\hline 
\text{\begin{tabular}[c]{@{}l@{}}\textbf{Ablation Study} \end{tabular}} &
\text{\begin{tabular}[c]{@{}l@{}}\textbf{Parameter} \end{tabular}} &
\text{\begin{tabular}[c]{@{}c@{}}Model\\ Inf\\ Time\\ (ms)\end{tabular}} &
\text{\begin{tabular}[c]{@{}c@{}}Model\\ Size\\ (MB)\end{tabular}} & \text{\begin{tabular}[c]{@{}c@{}}OPs\\ (10\textsuperscript{9})\end{tabular}} & \text{\begin{tabular}[c]{@{}c@{}}Segment\\ IOU\\ (\%) \end{tabular}} & \text{\begin{tabular}[c]{@{}c@{}}Pose\\ IOU\\ (\%) \end{tabular}} \\ \hline \hline
\multirow{2}{*}{\textbf{Convolution Type}} & \textbf{Depthwise Conv} & \textbf{4.7} & \textbf{1.7} & \textbf{1.5}  & 97.19 & 89.60  \\ \cline{2-7}
 & \textbf{Group Conv*} & \textbf{4.7} & 7.2 & 4.8 & \textbf{97.41} & \textbf{90.20} \\ \hline \hline
\multirow{3}{*}{\textbf{Width Multiplier}} & \textbf{\space\space1.0*} & \textbf{5.7} & \textbf{9.4} & \textbf{5.3} & 97.50 & 91.40  \\ \cline{2-7} 
 & \textbf{1.5} & 8.2 & 21.0 & 12.1 & 97.89 & 92.37 \\ \cline{2-7} 
 & \textbf{2.0} & 11.4 & 37.0 & 20.6 & \textbf{97.96} & \textbf{92.63} \\ \hline \hline
\multirow{3}{*}{\textbf{Decoder Type}} & \textbf{Bilinear Upsampling} & \textbf{6.5} & \textbf{8.7} & \textbf{7.4}  & 97.65 & 90.65  \\ \cline{2-7} 
 & \textbf{Channel Attention} & 8.2 & \textbf{8.7} & 7.5 & 97.75 & \textbf{91.81} \\ \cline{2-7} 
 & \textbf{Multilayer Perceptron*} & 6.9 & \textbf{8.7} & 7.6 & \textbf{97.81} & 91.80 \\ \hline \hline
\multirow{4}{*}{\textbf{Image Resolution}} & \textbf{256x256} & \textbf{1.6} & \textbf{8.7} & \textbf{1.9} & 96.94 & 87.23  \\ \cline{2-7} 
 & \textbf{384x384} & 3.1 & \textbf{8.7} & 4.3 & 97.51 & 90.10 \\ \cline{2-7} 
 & \textbf{\space\space512x512*} & 6.9 & \textbf{8.7} & 7.6 & 97.79 & 91.80 \\ \cline{2-7} 
 & \textbf{640x640} & 9.5 & \textbf{8.7} & 12.0 & \textbf{97.98} & \textbf{92.94} \\ \hline
\end{tabular}
\end{adjustbox}
\vspace{-2mm}
\caption{Ablation study over edge segmentation model design parameters. ``$\ast$'' denotes our final choice in each ablation category.}
\label{table:ablation}
\vspace{-4mm}
\end{table}

\section{E2E Heterogeneous Pipeline Optimization}
\label{sec:inference-pipeline-opt}

% When designing a video segmentation system for video conferencing or human video editing in AR,
One of the most important performance criteria of a video segmentation
system for AR applications is the E2E frame rate, which is directly perceivable by
the application users.  % and also characterizable by the system developer
% with the underlining inference latency and the end-to-end pipeline latency.
In this section, we discuss our particular considerations for ML inference
pipeline optimization on web browsers and mobile devices leveraging the GPU and the NPU, respectively.

\begin{figure*}[ht]
    \centering
    \includegraphics[width=1.0\textwidth]{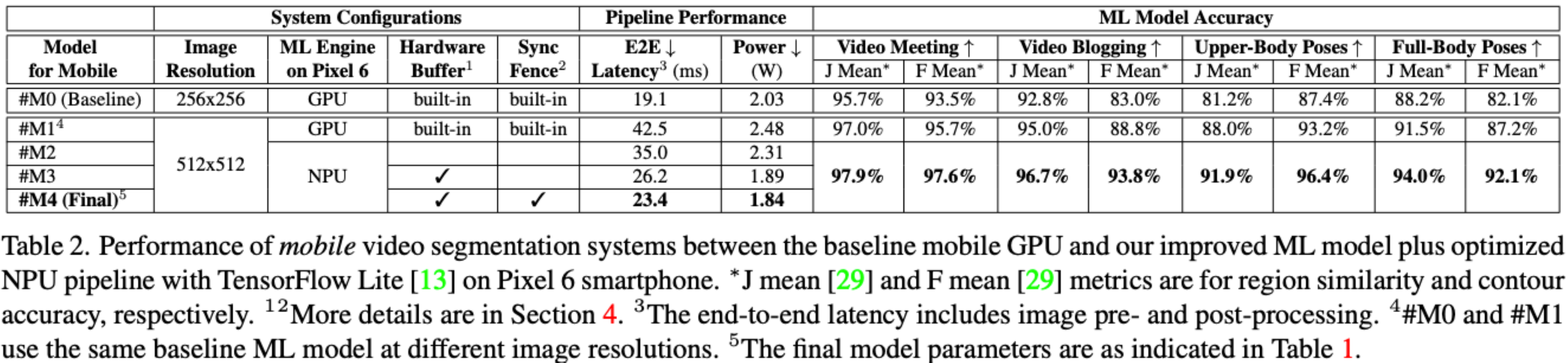}
    \label{fig:decoder}
\vspace{-5mm}
\end{figure*}

\textbf{GPU for the Web:}
In browsers, ML
frameworks with WebGL support~\cite{tfjs,ortweb,paddlejs} were our
natural choices for hardware-aided ML inference in the browser, as NPUs currently lack a web standard.
Since these implementations ran significantly slower than native
OpenGL~\cite{tflitegpu}, we wrote our own WebGL inference engine that can run
at near-native OpenGL performance.  In our WebGL implementation,
performance-critical operations,~\eg convolutions, leverage a modern
GPU feature called Multiple Render Targets (MRT)~\cite{mrt},
which allows rendering multiple textures at once, substantially
reducing the overhead of multiple draw calls.  For efficient MRT, we
separate logical tensors and physical GPU objects, which have a $1$-to-$1$
correspondence in other frameworks, and allow tensors to take flexible
layouts instead of one hard-coded layout.

\textbf{NPU for Mobile Devices:}
Advanced mobile-optimized video processing frameworks, such as
MediaPipe~\cite{mediapipe}, offer streamlined pipelines primarily targeting
image and video processing along with downstream co-processor interactions
and GPU execution, typically for the tasks of
image acquisition, pre-/post-processing, and rendering.
For NPU-accelerated ML inference use cases,
the system can benefit from a processing chain that is tightly
coupled with the NPU. A naive % adaptation of the processing pipeline by
replacement of GPU inference with its NPU counterpart can introduce major inefficiencies such as
indirect inter-processor data flow and
latency from inter-processor synchronization.
We streamline the former with Native Hardware Buffers~\cite{ahardwarebuffer}
supporting shared access by both the GPU and the NPU, and
reduce the latter with sync fences in the Android Synchronization API~\cite{sync_fence}.

\section{Results}
\label{sec:results}

Our segmentation models are trained with a standard Jaccard~\cite{jaccard, tanimoto1958elementary} loss on annotated datasets totaling 250k images that cover a wide range of video scenarios, from video conferencing to human activities. We evaluate our models on proprietary datasets,~\eg ``video meeting'' ($800$ samples) and ``full-body poses'' ($947$ samples), and report their mean Intersection over Union (mIoU)~\cite{iou}, J Mean~\cite{Perazzi_CVPR_2016} for region similarity, and F Mean~\cite{Perazzi_CVPR_2016} for contour accuracy.

\textbf{GPU for the Web:}
For video segmentation in web apps designed for desktop and laptop web browsers,
we run our EfficientNetLite- and MobileNetV3-based segmentation models with the
baseline CPU (XNNPACK~\cite{xnnpack} with WebAssembly SIMD~\cite{wasm}) and with the GPU (our
WebGL implementation).
Tab. \textcolor{red}{3} presents the direction for ML inference in the web
browser: (a) to employ the GPU for Web ML when possible, and (b) to prefer
EfficientNetLite over MobileNetV3 for GPU-accelerated video segmentation on the web.
In fact, the latter observation led to us confidently pursuing
EfficientNetLite-based models for the NPU below.

\begin{figure}[t]
\centering
\includegraphics[width=\columnwidth]{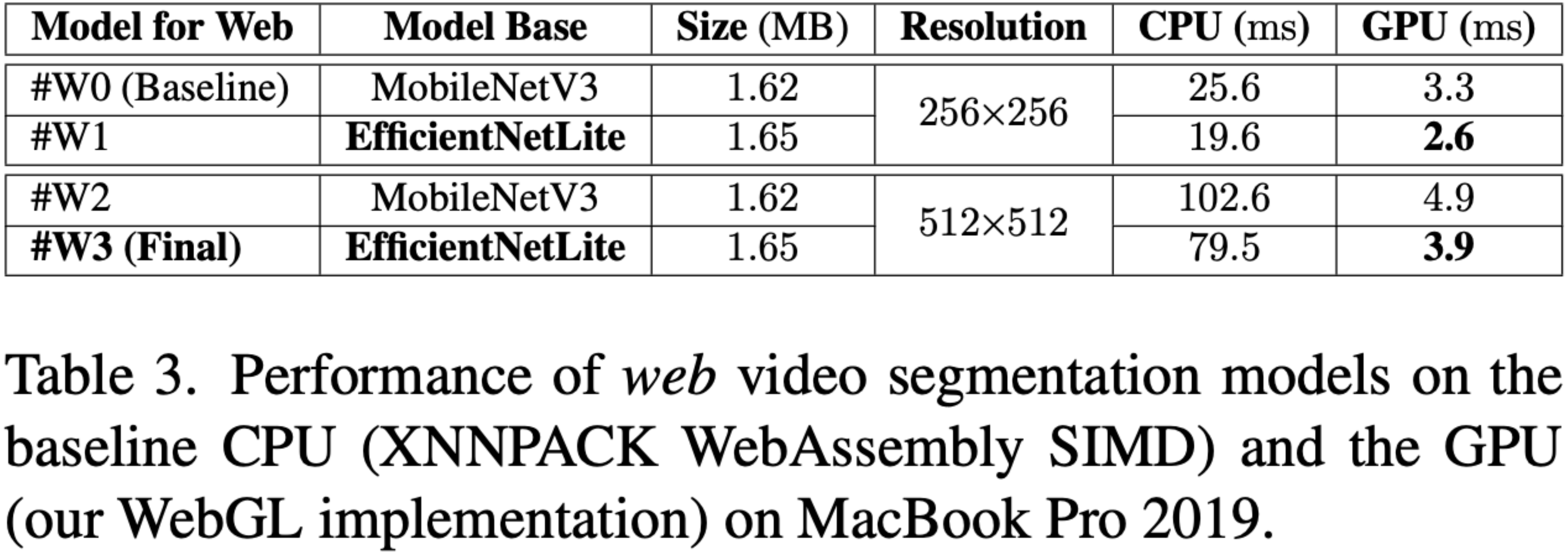}
\vspace{-7mm}
\end{figure}

\textbf{NPU for Mobile Devices:}
% Additionally, for the end-to-end NPU pipeline on smartphones,
With our model improvement and pipeline optimization as discussed in Sections \ref{sec:model} and \ref{sec:inference-pipeline-opt},
we have achieved significantly higher accuracy and enabled much shorter inference latency, while reducing power consumption compared to the baseline.
For end-to-end performance evaluation, we run video segmentation tasks with 512x512 input image resolution and 7.6B FLOPS in various architectures on a Google Pixel 6 smartphone.
As shown in Tab. \textcolor{red}{2}, the Model System \#M4 (Final) with configurations of the NPU, shared data buffer, and asynchronous GPU-NPU execution along with other choices indicated in the ablation study of \cref{sec:model},
% Table \ref{table:ablation} (ablation study),
outperforms \textit{all} other variants in \textit{both} latency and power consumption. Remarkably, our final model \#M4 achieves higher accuracy with a speed up by $81\%$ (at $42.7$ inf/sec) while consuming only $74\%$ power of \#M1 with the baseline ML model as shown in Tab. \textcolor{red}{2}.

\section{Conclusion}
\label{sec:conclusion}

In this paper, we showcase a development for edge-based video segmentation with our improvements and optimizations of ML model, inference acceleration and E2E heterogeneous compute pipeline.  We visit our deliberations that cover the scope of model architecture design and system pipeline analysis, targeting specifically resource-limited edge ML.  We present our design and optimization methodology in the ML model architecture and in the end-to-end ML pipeline with efficient use of resources and APIs to maximize the throughput for production-ready, heterogeneous edge ML solutions.  We demonstrate that our methodology enables higher accuracy across various datasets, lower end-to-end inference latency, and lower power consumption.  We expect our methodology to be beneficial and extensible to a wider range of edge CV/ML systems and real-time AR/VR applications.

\newpage

%%%%%%%%% REFERENCES
{\small
\bibliographystyle{ieee_fullname}
\bibliography{egbib}
}

\end{document}